\documentclass{article}
\usepackage[utf8]{inputenc}

\usepackage{float}
\usepackage{hyperref}
\usepackage{todonotes}

\usepackage[style=ieee]{biblatex}
\usepackage{color, colortbl}
\definecolor{nsig}{rgb}{1,0.7,0.7}
\definecolor{sig}{rgb}{0.7,1,0.7}

\usepackage{geometry}
\title{Inferring Intentions to Speak Using Accelerometer Data In-the-Wild}
\author{Litian Li\footnote{L.Li-35@student.tudelft.nl}, Jord Molhoek\footnote{j.molhoek@student.tudelft.nl}, Jing Zhou\footnote{j.zhou-15@student.tudelft.nl}\\Delft University of Technology, Intelligent Systems Department\\The Netherlands}
\date{January 2023}

\begin{document}

\maketitle
\begin{abstract}
% the abstract should summarize the contents of the paper and should contain at least 70 and at most 150 words. The paper itself is max 10 pages in this format

Humans have good natural intuition to recognize when another person has something to say. It would be interesting if an AI can also recognize intentions to speak. Especially in scenarios when an AI is guiding a group discussion, this can be a useful skill. This work studies the inference of successful and unsuccessful intentions to speak from accelerometer data. This is chosen because it is privacy-preserving and feasible for in-the-wild settings since it can be placed in a smart badge. Data from a real-life social networking event is used to train a machine-learning model that aims to infer intentions to speak. A subset of unsuccessful intention-to-speak cases in the data is annotated. The model is trained on the successful intentions to speak and evaluated on both the successful and unsuccessful cases. In conclusion, there is useful information in accelerometer data, but not enough to reliably capture intentions to speak. For example, posture shifts are correlated with intentions to speak, but people also often shift posture without having an intention to speak, or have an intention to speak without shifting their posture. More modalities are likely needed to reliably infer intentions to speak.

% Why? motivation, relevance, scientific & application
% Make machines able to recognize intentions to speak. Can be useful for e.g. AIs that lead group discussions.

% What? The concrete research challenge or question
% Recognize intentions to speak in and in-the-wild setting. Using accelerometer data since it is privacy-preserving and feasible for in-the-wild setting since it can be placed in a smart badge

% How? The approach to address that challenge, incl evaluation
% Implement exploratory study of the dataset, manually annotate those perceived unsuccessful intentions to speak, use existing model to train with successful intentions and test with manully annotated unsuccessful intentions to speak, evaluate the model quantitatively.

% Results? The outcomes and the contributions
% There ís useful information in accelerometer data, but not enough to reliably capture intentions to speak. E.g. posture shifts are correlated with intentions to speak, but people also often shift posture without having an intention to speak, or have an intention to speak without shifting their posture. More modalities are needed. Perhaps audio modality.

% Abstract serves as a short summary, used to enhance finding the paper and for advertising it. And reviewing.

\end{abstract}

\section{Introduction and Motivation}
% Here you introduce the target paper’s topic, its relevance, how it fits in the field of social signal processing. The refs you use here are to give context and understanding to the reader and to “carve out” the topic of the target paper.
Being able to automatically infer intentions to speak can be applied in many social scenarios. For example, a social robot that leads a group conversation that recognizes when someone had something to say but did not get the chance to speak up can lead the turn to this person. This can potentially make the group conversation more effective by having more people share their thoughts and contribute to the conversation. 
%In a natural unorganised conversation, the turns are often managed by the most dominant participant\todo{reference}. 
Inferring the intentions to speak can also have a positive effect on those with a tendency to be shy or less dominant, which helps them to realize the unsuccessful intentions to speak, and makes them feel more engaged and involved in the group. Besides, at the group level, it also helps with better group dynamics and equality by giving every participant an equal chance to speak up and share. Furthermore, almost any social robot in any social situation can be improved by being equipped with the skill of recognizing intentions to speak, because it can always happen that turns collide and people that have something to say do not get a chance to speak. In any situation, a robot that is able to naturally say ``Excuse me, I thought you had something to say just now. Could you please share that with me/us?" can be perceived as a more natural and/or enjoyable conversation participant. Such applications are of course not limited to social robots but can be any smart interface which leverages such functionalities. Finally, we hope to contribute to the overall progress of the field of social signal processing - with the ultimate goal of having social robots and agents that can recognize and synthesize more social signals.

Next speaker prediction is different from speaking intention prediction in the sense that also the unsuccessful intention cases - those where there was an intention to speak, but the conversation participant did not end up getting the turn - also need to be recognized. It is difficult to accomplish the task of predicting the intention, since the intention itself is based on the psychological thought, instead of the ground truth of an actual action. Such thought might not even be manifested through any social cue and hence not be observable. However, as indicated by Petukhova and Bunt \cite{whosnext}, the two fields are closely related. An important difference is that next speakers can be predicted using features of group dynamics. An example of this is when all participants gaze at one person, this person might feel pressured to say something \cite{weiss2018gaze}. These features based on the group dynamic are not useful for inferring intentions to speak, because the features that could indicate an intention to speak are purely individual. Currently, few techniques are focusing on predicting the intentions compared to those predicting the actual action. Wlodarczak and Heldner \cite{wlodarczak2020breathing} argue that existing techniques exploiting respiration to infer speaking intentions are still far from clear. Existing methods for predicting intentions to speak are mainly lab-based, such as \cite{respiration_with_sensors, mouth_open_patterns}. Hence, it is worth filling in the gap between inferring the intentions to speak and next-speaker prediction. The aim is to do this in a way that works in-the-wild (in other words; out of the lab).

In this work, we split up intentions to speak into successful cases (where there was an intention to speak and the participant took the turn) and unsuccessful cases (where there was an intention to speak and the participant did not get the turn). The unsuccessful cases can be further split up into perceived/perceivable intentions to speak and unperceived/unperceivable intentions to speak. The latter are intentions that are not manifested through any social cue or physiological feature; these cases will be extremely hard - arguably impossible - to recognize, and are therefore outside the scope of this work.

Some methods for next speaker prediction exist, such as \cite{respiration_with_sensors, mouth_open_patterns, ishii2016prediction, ishii2015multimodal, ishii2015predicting, kawahara2012prediction, malik2020speaks}. The goal of this research is to use existing methods for next speaker prediction for inferring intentions to speak in-the-wild.

\subsection{Related work}
% Here you present the actual review of your target paper’s topic. Don’t attack it, review the topic and the findings in the paper critically on their up- and downsides. Also include related work that you need in order to have the reader understand your review and point of view. This is in fact the typical “motivation and related work” section in a paper as this section should critically describe the work of others and how those works relate to your project. Issues to be covered include all stuff related to eventually understanding why you propose the question you want to address in section 2.
Since our task is to infer intentions to speak, it can be related to three types of existing research: turn-taking (in a multiparty conversation), next-speaker prediction, and the intention of an individual to speak. In this section, we review some related work of these three aspects respectively. However, based on our review, we find that current work mostly focuses on turn-management prediction and next-speaker prediction. There are few existing techniques aiming to infer the intentions to speak, and our project aims to focus on the gap between predicting the action and the intentions. In our case, it is the gap between predicting the actual next speaker and inferring any possible participant who has the intention to speak.

\subsubsection{Turn Taking}
According to Petukhova and Brunt \cite{whosnext} turn management is an essential aspect of any interactive conversation and involves highly complex mechanisms and phenomena. There already exists various literature that investigates turn-taking in conversations. 

Sacks et al. attempts to characterize the organization of turn-taking in the simplest form by using audio recordings of naturally occurring conversations \cite{sacks1978simplest}, Some of the observations are: it is common that more than one speaker at a time, but the time interval is brief; transitions with no gap and overlap are common, together with transitions with slight gap or overlap, they make up the majority of transitions; turn-allocation techniques are used; repair mechanisms exist for dealing with turn-taking errors and violations; if two parties find themselves talking at the same time, one of them will stop prematurely \cite{sacks1978simplest}.

The work of Petukhova and Brunt \cite{whosnext} investigates how some social cues are correlated with receiving the next turn. These cues are called turn-initial signals. This research found strong correlations with gaze aversions, lip movements and posture shifts. Specifically, 10.9\% of the turns taken were preceded by a lip movement (compressing, biting or licking). 47.3\% of the taken turns were preceded by a (half-)open mouth. Mouth-opening patterns as turn-initial signals are further discussed in \cite{mouth_open_patterns}. Petukhova and Brunt \cite{whosnext} also found that fillers such as ``um" and ``right" often occurred right before someone took a turn. 
As mentioned before, gaze is also used as a turn-initial signal. According to Novick et al. \cite{novick_gaze} 42\% of the turn changes follow the pattern: 1. the speaker looks towards the listener as they complete the turn; 2. the speaker and the listener have a short moment of eye contact; 3. the listener looks away and begins speaking. Automatic recognition of such patterns can be useful for next speaker prediction and possibly for prediction of intentions to speak. Petukhova and Brunt \cite{whosnext} also found that ``Participants who used more than one turn-initial signal or two modalities were more successful in obtaining the next turn." Hence, basing a prediction on just one feature will not be enough. The key is in combining the different turn-initial signals/modalities.

Moreover, Auer \cite{auer2021turn} shows that the gaze of both the speaker and listeners on the last part of a speaker's turn are very powerful features to predict the next speaker in a multiparty setting, and it is the most ubiquitous one among various techniques. Closely related to this, Weiß \cite{weiss2018gaze} shows how gaze behaviour influences the next speaker selection or self-selection in multiparty scenarios. The findings show that speaker gaze can elicit a response from a recipient, and if the speaker is looking at the recipient, the recipient is more likely to respond. However, if the recipient is also looking at the speaker, the recipient is more likely to respond promptly without delays, and looking at the recipient can exert additional pressure and display that a response would be expected by the end of the turn. Furthermore, it illustrates that a speaker who is gaze-selected by the previous speaker can either pass on the turn to a third participant by gazing at them, or they can reject the offered turn by ‘gazing away’ and dissolving the mutual gaze, and thus opening up the conversational floor for the other participant to self-select. 

In a multi-party conversation, turn-taking can be generated smoothly by sending and receiving social cues among all participants, but sometimes can also become rigid, as an interruption, which can be interpreted as ``not letting current speaker finish" \cite{schegloff2001accounts}. Interruptions in a conversation might lead to different perceptions about participants' awareness as well as emotions. In \cite{cafaro2016effects}, interruptions are divided into two categories: cooperative and intrusive, the former refers to the kind of interruptions intended to coordinate with the current speaker, whereas the latter refers to the ones that disrupt the current process of the conversation. Researchers argue that different types and strategies of interruptions could lead to varying degrees of perceived interpersonal attitude, a cooperative strategy leads to more engagement and involvement in the interaction, while the intrusive interruption works in the opposite way \cite{cafaro2016effects}. 

Thorisson \cite{thorisson2002natural} presents a model that is based on the hypotheses of turn-taking mechanisms, and formalises the elements needed for a complete generative model of real-time, face-to-face turn-taking. This includes the modalities of speech, prosody, body language, manual gesture, and gaze as input, and generates multi-modal output. 

The Ymir Turn Taking Model (YTTM) is an agent-oriented model, motivated by a cognitive focus, it addresses the full perception-action loop of real-time turn-taking, from (1) the basics of multi-modal perception to (2) knowledge representation, (3) decision making, and (4) action generation for gaze, gesture, facial expressions and speech planning and execution \cite{thorisson2002natural}. It has been implemented in numerous interactive systems and has been extended in various ways. Thórisson et al. \cite{thorisson2010multiparty} investigate how the YTTM accommodates multi-party dialogue, and extend the YTTM model with a test of up to 12 agents in a virtual world. The extension of Thórisson et al. \cite{thorisson2010multiparty} implements 8 dialogue contexts, with each containing different perception and action modules, representing the disposition of the system at any point in time. Additionally, the results of \cite{jonsdottir2009teaching, thorisson1999mind, thorisson2010multiparty} show that YTTM 
possesses the ability to address real-world and real-time dialogue, though future work needs to extend the content generation, and interpretation mechanisms \cite{thorisson2010multiparty}.

Blythe et al. \cite{blythe2018tools} use data from remote Aboriginal communities for research on next speaker selection mechanisms and avoiding being selected in multiparty conversation by focusing on the features of gaze direction, bodily orientation and pointing. They also investigated the effect of some verbal features such as voice projection, interrogative grammar, prosody, and other features like epistemic status.

\subsubsection{Human Intentions}
% \textbf{[Psychology about intentions to speak. Look at Marcin Wlodarczak papers. Also look at work by Bratman https://www.jstor.org/stable/pdf/2215590.pdf about relationship between intentions and actions]}

Some social science literature distinguishing people's intentions and actions is helpful to understand the difference between the intention to speak and the real action of speaking. This could help to transform current next-speaker prediction techniques into potential intention-inferring models. 

Blakemore and Decety \cite{blakemore2001perception} argue that humans have an inherent tendency to infer other people's intentions from their actions. The mechanism for inferring intentions from observed actions may be based on simulating the observed action and estimating an actor’s intentions based on one’s own intentions \cite{blakemore2001perception}, which means we as humans already have the ability to infer other people's intentions. Understanding how humans infer others' intentions could contribute to a model that inferences a human's intentions more accurately. Furthermore, this also hints at the possibility of a model that exploits the help of our intuition. 

Bratman studied intention methodologically, his work proposed that an intention can be understood as a manifestation for humans to plan and coordinate activities in advance \cite{bratman1987intention}. Bratman argues the main role of intention is to commit an agent to an action, and this commitment can be divided into two aspects: reasoning-centred commitment and volitional commitment, in which the first one enables one to deliberate to lay a stable groundwork on which future deliberations can build, while the second commitment controls action rather than merely influences it \cite{bratman1987intention}. The theory of Bratman concludes that intending to do something does not entail that one will do it eventually \cite{bratman1987intention}, which supports the common sense that in a multi-party conversation an intention to speak does not always result in successful turn-taking.

Wlodarczak and Heldner \cite{wlodarczak2020breathing} have done the first study that attempts to identify unrealized turn-taking intentions by a fully automatic method. As the authors indicate, breath holds are sometimes produced toward the beginning of silent exhalations, potentially indicating an unrealised intention to turn-taking, thus the breathing signal can be used for hidden turn-taking events \cite{wlodarczak2020breathing}.

\subsubsection{Next Speaker Prediction Techniques}
In the task of predicting the next speaker, there exists work on predicting the next speaker based on verbal and non-verbal features. Based on these techniques, it may be possible to get some inspiration for detecting intentions to speak. Some highlights are discussed in this section. 

% NSP with Respiration features CAN STAY HERE
By looking at non-verbal features only, Ishii et al. \cite{respiration_with_sensors} investigate a three-step model using participants’ respiration to predict the next speaker. This model first predicts if a turn is kept or changes to another speaker. If it is a turn-change, then the second step is to predict the next speaker, and the last step is to predict the next-utterance timing. Meanwhile, if it is a turn-keep the model then goes directly to the third step, which is to predict next-utterance timing \cite{respiration_with_sensors}. The model of \cite{respiration_with_sensors} is being tested for prediction 200ms after the end of utterance to clarify whether respiration is a useful feature in this case. Based on the prediction results, researchers found several rules based on the respiration changes for the parties who participated in a multiparty meeting. For example, a speaker takes a breath more quickly and deeply after the end of an utterance in turn-keeping than in turn-changing \cite{respiration_with_sensors}. Likewise, the listener who will be the next speaker takes a bigger breath more quickly and deeply in turn-changing than the other listeners \cite{respiration_with_sensors}. As Petukhova and Bunt \cite{whosnext} indicated, Ishii et al. \cite{respiration_with_sensors} also argue that using respiration as the only feature to predict the next speaker could be improved by a multi-modal processing model that takes gaze and head movement into account. The depth and speed of breathing may also reflect the degree of intention to speak for an individual, and successful and unsuccessful speaking intentions may have similar respiration patterns. This can be a useful feature to infer intentions to speak.

% Mouth-Opening Transition Pattern CAN STAY HERE
In the study of Ishii et al. \cite{mouth_open_patterns}, Mouth-Opening Transition Patterns (MOTP) are also used for the prediction of who will be the next speaker in a multiparty conversation. Their model uses the change of mouth-opening degree during the end of an utterance to predict the next speaker, and the interval between the start time of the next speaker's utterance and the end time of the current speaker's utterance. The data collected includes verbal data to identify the start and the end of the utterance, and non-verbal data which is the degree of mouth opening data. There are three categories for the different degrees of mouth opening; closed, narrow-open and wide-open. The results show that for predicting the next turn the current speaker often keeps their mouth narrow-open during turn-keeping and starts to close it after opening it narrowly or continues to open it widely during turn-changing. Meanwhile, the next speaker often starts to open their mouth narrowly after closing it during turn-changing. For predicting the utterance interval, the current speaker starts to close their mouth after opening it narrowly in turn-keeping, the utterance interval tends to be short, and when the current speaker and the listeners open their mouths narrowly after opening them narrowly and then widely, the utterance interval tends to be long. In conclusion, making use of both MOTPs and eye-gaze behaviour in a multimodal model can improve effectiveness for predicting the next speaker. Furthermore, people have a tendency to open their mouths slightly before they start speaking. Perhaps this idea is not only a turn-initial signal but also relates to intentions to speak.

% Gaze-related features 
Closely related to the work of Auer \cite{auer2021turn} and Wei{\ss} \cite{weiss2018gaze} as discussed before is the work of Ishii et al. \cite{ishii2016prediction}, which investigates using gaze behaviour to predict the following three tasks in multiparty meetings. The first is whether turn-changing or turn-keeping occurs, the second is who will be the next speaker in turn-changing and the third is the timing of the start of the next speaker's utterance. The gaze behaviour can be subdivided into two features; the gaze-transition patterns providing information about the order in which gaze behaviour changes and the time structure of eye contact between a speaker and listener representing who looks at whom first and who looks away first. The data collected was the 1000ms window before the end of IPUS (end of utterance) and 200ms after IPUS; a total window of 1200ms. The result showed that Gaze-transition patterns are useful for all three tasks, and the timing structure of eye contact was not useful to predict tasks two and three. Combining the timing structure of eye contact with gaze-transition patterns can improve performance for predicting turn-changing and turn-keeping. Although it is not investigated directly by Ishii et al., gaze can still be a promising feature for inferring intentions to speak; considering that the person with an intention to speak may gaze at the current speaker expectantly.

% MODEL for the fusion of gaze and breath; can stay here
Ishii et al. \cite{ishii2015multimodal} also investigate using both respiration and gaze for multimodal fusion to predict the next speaker. The prediction should be more robust and high-precision by using multimodal information. Respiration conveys information about the intention or preliminary action to start to speak, while gaze behaviour reveals the information of transferring the speaking turn between humans. The results suggest that the model with both respiration and gaze behaviour performs better than using only one of them. Additionally, it was found that respiration is more useful for predicting the next speaker in turn-taking than gaze behaviour in this fusion model. 

% MODEL for NSP from head movement; can stay here
Other than that, the relationship between head movement and the next speakers in multiparty conversations has also been investigated by Ishii et al \cite{ishii2015predicting}. The features of head movement are identified as the values of the position of the head (X, Y, Z) and the rotation of the head (azimuth, elevation, roll) for all participants. The results revealed listener's head movement can contribute to predicting the next speaker in turn-taking. 

% \textbf{[removed speaking status]}

% \todo{refactor this speculation}
% For the head movement and accelerometers described above, they may also reflect the individual intentions to speak, although this is not investigated in detail. For example, at the moment that a person intends to grab the turn to speak, there may be drastic changes in head motion or posture shifts over a short period.

When the verbal features are included as well, the work of Kawahara et al. \cite{kawahara2012prediction} investigates using both backchannels and eye-gaze features to predict the next speaker in multiparty conversation. Verbal and nonverbal backchannels - such as ``yeah", ``okay" or nodding - indicate that the listener understands what is being said and is active in the interaction. Moreover, the eye-gaze indicates the eye-gaze object and joint eye-gaze event, and their duration. The findings show that using a combination of all of them performs significantly better than using only eye-gaze or backchannels. 

% Dialogue-related features
Other features have also been used to predict the next speaker by Malik et al. \cite{malik2020speaks}, such as speaker role during multiparty interaction, dialogue act, pause duration, start and end time of utterance, the focus of attention and addressee role. By applying an ablation study, it was found that both pause duration and addressee role played important roles in improving performance.

\subsubsection{Speaking Status Inference}
Similar to, but different from next speaker prediction is speaking status inference. To detect the speaking status, Vargas and Hung developed a no-audio multi-modal approach by using body-worn accelerometers and video only \cite{quiros2019cnns}. The visual modality allows more detailed computational analysis of social behaviour, while the accelerometers in wearable devices pose little privacy concerns as it only provides information about the general body movement of the Participants.

\section{Research question}
\label{sec:rq}
% Here you detail the research question, and how this relate to the work of others. What makes this research novel and interesting. Why is it worth studying. Remember, a research question is a real question!?.
As discussed before, methods for next-speaker prediction already exist. Some methods to infer intentions to speak also exist, but only in lab settings. It is interesting to find a privacy-preserving and ubiquitous technique to infer intentions to speak in a real-life setting; accelerometers would be a good candidate as they collect data in a privacy preserving way and can be built in a smart badge. Hence, the main research question of this work is: \textbf{to what extent can accelerometer data be used to predict intentions to speak in-the-wild?}

Interesting sub-questions are: 
\begin{enumerate}
    \item To what extent can accelerometer data be used to predict successful intentions to speak?
    \item To what extent can accelerometer data be used to predict unsuccessful intentions to speak?
\end{enumerate}

Note that in this work we define ``intentions to speak" as intentions to take the turn and make an utterance. Hence, back-channels are excluded.

\subsection{Hypotheses} 
\label{sec:hypotheses}
% Here you state your hypotheses. A hypothesis is a testable yes/no question derived from your research question. For example: nice weather increases the number of incidents on highways.

% If a person takes a turn, this person had an intention to speak. This idea indicates a strong correlation between intentions to speak and next speakers. Perhaps this idea is incomplete because there are also unsuccessful intention-to-speak cases. In these unsuccessful cases, the participant likely had the same or similar behavioural patterns compared to the successful cases. This is something that can potentially be exploited. Therefore we hypothesize that existing methods for next speaker prediction can be used to predict successful intentions to speak. Furthermore, we hypothesize that existing methods for next speaker prediction can also be used to predict unsuccessful (but perceived/perceivable) intentions to speak, but with lower precision and/or recall than the successful cases. Possibly some novel extensions are needed for inferring intentions with higher precision and recall.
To investigate the research sub-questions listed in section \ref{sec:rq}, the two following hypotheses are formulated:
\begin{enumerate}
    \item A model trained with only accelerometer data and successful intentions to speak as ground truth can predict successful intentions to speak better than random guessing.
    \item A model trained with only accelerometer data and successful intentions to speak as ground truth can predict unsuccessful intentions to speak better than random guessing.
\end{enumerate}

Random guessing is chosen as the baseline because the goal is to see if accelerometer data contains useful information for the purpose of inferring intentions to speak. Section \ref{sec:future} discusses how models could be improved and higher model quality might be achieved.

\section{Method}
% You describe how you study your hypotheses, what kind of approach you took (e.g., an experiment). If you write the proposal, you will have to write section 3.4, in the final paper that is left out.
To truly predict intentions to speak, one would need interaction data that is self-annotated by the participants where they mention exactly when they had an intention to speak. These would be the truly correct ground truth labels, although one could still argue that these annotations are still not 100\% correct because the people might have forgotten some cases or are maybe more likely to remember the successful cases. We decided to limit our scope to inferring perceived/perceivable intentions to speak. This opens the door to third-party annotations of perceived intentions to speak. Limiting our scope in this way will result in some missed cases, where a participant had an unsuccessful intention to speak, and did not make this perceivable in any way. This is a sacrifice we are willing to make because inferring the intentions to speak that are not manifested through any social cue would be infeasible (arguably impossible) within the frame of time and resources for this research.

Furthermore, we desire to infer (perceivable) intentions to speak in a privacy-preserving and ubiquitous way. We do this inference based on accelerometer data only because accelerometers capture a superposition of useful features for intention to speak prediction in a privacy-preserving and ubiquitous way. Examples of such features are body movement, posture shifts and possibly breath patterns, although this final feature will likely not be captured in fine-grained detail, as a wearable accelerometer such as a smart badge will not be as sensitive as a specialized breath sensor. This is a trade-off because these specialized breath sensors, such as those used in \cite{respiration_with_sensors} are not feasible in-the-wild.

\subsection{Materials} 
% You describe the material you used (i.e., the system you developed, the experimental materials used to test the system)
We have considered four datasets for this research. Namely, Conflab \cite{conflab, conflab-code}, REWIND\footnote{Obtained through personal communication with the authors.} \cite{laughterquiros2023}, Dominos\footnote{Placeholder name for an unpublished dataset by the University of Hamburg} and MULAI \cite{mulai}. They are compared and described in detail in the appendix \ref{data-considerations}. The final decision was in favour of the REWIND dataset. This dataset is described in more detail in this section.

The REWIND dataset is chosen as the main dataset for this research because REWIND collects the data of an indoor social networking event when standing people are free to walk and talk to others. Video data is recorded by overhead cameras and elevated side-view cameras. Additionally, a subset of individuals has wearable accelerometers. A subset of participants also has wearable microphones. The intersection of participants with a microphone and an accelerometer consists of 24 people. In some segments, the crowd is collectively listening to music or a public speaker. There are also segments where participants are assigned a conversation partner and a topic/question to discuss with them. Sifting out these segments, roughly one hour and fifty minutes of in-the-wild free social networking data remain.

\subsection{Experimental Approach}
% You describe how the experiment was setup (is applicable). 
As implied by the research sub-questions, we split up our approach into two cases: inferring successful intentions to speak and inferring unsuccessful intentions to speak. Therefore, we first fully focus on inferring successful intentions to speak (from the accelerometer data of an individual). Originally, we aimed to build a model that can infer successful intentions based on existing next-speaker prediction techniques. Since none of the techniques we could find was privacy-preserving, ubiquitous and could work in-the-wild, we decided to refactor existing code for speaking status inference and only use the accelerometer data. The workings of this model are explained in \ref{sec:model}. 

We made one important simplifying assumption about the modelling approach; we approach it as a classification problem. 
It could be argued that an intention to speak is continuous, as sometimes an intention to speak is not as strong as another time. However, we assume that intentions to speak are binary; someone wants to say something, or they do not.

With the model at hand, we inspect the false positives (those cases where a speaking turn is predicted, but the participant does not speak) to see if they - or a subset of them - are perhaps unsuccessful intentions to speak. For this, we annotate a subset of the data with perceived unsuccessful intentions to speak. 
This allows us to qualitatively analyse the model on both successful intentions to speak as well as perceived unsuccessful intentions to speak. The main measure we use for this evaluation is the AUC, which is discussed in section \ref{sec:measures}. Furthermore, the results are presented in section \ref{sec:res}.

\subsubsection{Exploratory Study of REWIND Dataset}
After looking at our own experiences and intuitions and after performing the literature survey, an exploratory study of the dataset was done in order to find even more social cues or features that can be useful for inferring intentions to speak. This was simply done by looking at and listening to randomly selected people in the REWIND dataset for a while, looking for possibly useful cues.

For this exploratory study, one author received access to a sound-proof booth in the INSYGHTLab\footnote{\url{https://www.tudelft.nl/ewi/onderzoek/faciliteiten/insyghtlab}} on the campus of TU Delft. With this, outside noise was ruled out completely, and it was possible to listen as closely as possible. Although it is good that it was tried, not a big difference between the booth and a silent private room using in-ear headphones was found.

\subsubsection{Successful Case Extraction}
\label{sec:successful__case_extraction}
To automatically get the successful cases (where a person actually gets to speak), we make use of microphone activation. A few problems with this approach are (1) microphone activation due to noise/other people speaking, (2) microphone activation because of short backchannels and (3) microphone deactivation when a speaker has a short pause but keeps the turn. Fortunately, the REWIND dataset solved the first problem for us, since diarized binary \textit{VAD} (Voice Activation Detection) files are provided for each participant that wears a microphone. The other two problems are solved by preprocessing. First, pauses shorter than 1.5 seconds are set to 1 (i.e. ``speaking"). Next, turns shorter than 1.5 seconds are set to 0 (i.e. ``not speaking"). These numbers are optimized empirically in a manual fashion.

\begin{figure}[]
\centering
\includegraphics[width=0.8\textwidth]{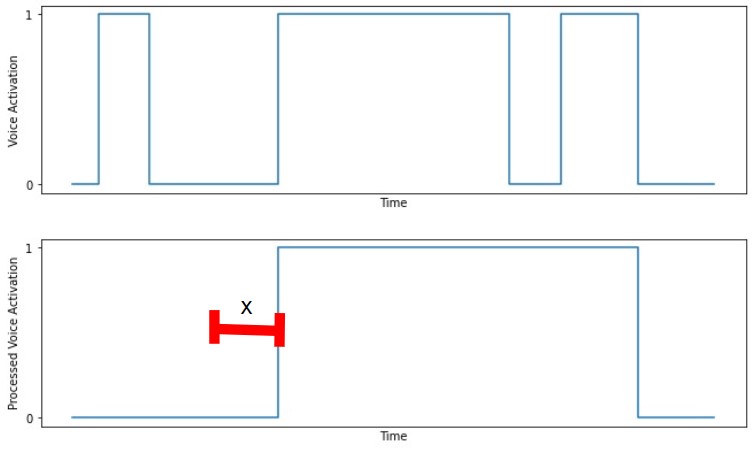}
\caption{Extraction of successful intention-to-speak cases.}
\label{fig:vad}
\end{figure}

After the VAD files are processed, the time windows with intentions to start speaking are extracted. This is done by finding the moments where speaking status jumps from 0 to 1 and extracting the $x$ seconds until that moment. This procedure is visualized in figure \ref{fig:vad}.

\subsubsection{Annotation}
In order to assess the quality of the algorithm on unsuccessful intentions to speak, a sample of unsuccessful intentions to speak is needed. This sample is generated by annotating a 10-minute segment of the REWIND dataset. A segment of 10 minutes was chosen because more would not be very feasible given the time and resource limitations of this project, and this 10-minute window would need to be multiplied by the number of (relevant) participants in order to get the total annotated time. The segment was chosen to be 1:00:00 until 1:10:00. This is somewhat arbitrary, but it does fulfil the requirement that the participants can move freely and talk to whom they want and about any topic they desire (e.g. the participants are not collectively listening to a public speaker and are not asked to play a conversational game with assigned conversation partners). Taking the intersection of participants with a microphone and an accelerometer, and removing participants that are not in view of any camera in the segment, 14 relevant participants remain. There was one person with audio issues, so in the end there were 13 participants of interest.

Given this segment, one (native Dutch) of the authors of this work performed the annotation. For this, the elan software was used \cite{elan}. The annotator listened to one participant at a time while looking at them through the camera in which they are best visible. When the annotator perceived a (likely) unrealized intention to speak, this segment was annotated. For completeness, two categories for intentions to speak were used: intentions to take the turn (labeled ``INTS\_start") and intentions to continue speaking (labeled ``INTS\_continue"). For these categories, 22 and 19 cases were found and annotated respectively.
Furthermore, for every annotation of a perceived unrealized intention to speak, there are annotations of the corresponding cues that lead to perceiving such an intention. These cues are ``posture change", ``audible smack/tongue click", ``filler", ``first word of utterance" and ``audible deep breath". After the annotations were finished, they were processed to a usable format in Python using the pympi library \cite{pympi-1.70}. 

In a perfect world, more data would have been annotated and multiple annotators would have been used. Then, the inter-annotator-agreement \cite{Artstein2017} would be used to assess the reliability of the annotations. Unfortunately, this was not feasible for this project.

\subsubsection{The Model}
\label{sec:model}
To answer the research questions and test the hypotheses, we refactored existing code\footnote{Obtained through personal communication with the authors of \cite{laughterquiros2023}.} in which three modalities: accelerometer data, audio and video are fused as input to a residual neural network model. This model was chosen because it is successfully used for speaking status inference. The model consists of three convolution layers, the kernel size of each layer is 3, 5, and 7 respectively. Since our project aims to be privacy-preserving, we cannot use raw data directly from audio and video. However, the accelerometer data meets both the requirement of in-the-wild and privacy-preserving, so we refactored the model by only using the accelerometer data as input of the residual neural network model. The code of our refactored model is shown in \cite{refactored-code}.

The model of the baseline existing code aims to predict speaking status instead of the intention of speaking, while our research topic is about intentions to speak, instead of the actual behaviour of speaking. Since the intention always happens before the actual speaking action, using the same data as in the speaking status prediction does not make sense. We choose to use the data from those time intervals that is \textit{X} seconds before the ground truth of speaking. We set the time interval \textit{X} as a parameter that can be changed to find the optimal time interval for performance, the results of different time intervals will be discussed in \hyperref[sec:quantitative]{4.2}. 

The Residual neural network is trained by 3-fold cross-validation and the batch size is 32. The number of epochs is decided after an experiment, which evaluates if a different number of epochs will affect the final result significantly. To train the model, we use data with successful intentions to speak, which is the accelerometer data when the participant is actually speaking. These training data are cut with the same time duration. For example, if the (processed) VAD file shows participant 1 is speaking from second 1800 until 1810, then the data we labelled as positive is the accelerometer data from 1800-\textit{X} to 1800, and the other data without the ground truth of speaking are labelled as negative samples.

% \todo{[JING] Write section about our model + Jose's code we used + how we refactored it}

\subsection{Measures}
\label{sec:measures}
% You probably have to measure something in order to test you hypothesis. This is where you explain what and how you measure this. For example, nice weather is measured as the average temperature over the day and incidents are all by the police reported traffic incidents on registered highways.

To evaluate the model, the AUC score is used. This stands for the area under the Receiver Operator Characteristic (ROC) curve, which is a performance metric for classification models. The ROC curve has the true-positive rate on the y-axis and the false-positive rate on the x-axis; representing the trade-off between these two. Since both the x-axis and the y-axis range from 0 to 1, the AUC also ranges from 0 to 1. 

In a ROC curve, a perfect classifier is represented as a vertical line from $(0,0)$ to $(0,1)$ and then a horizontal line from $(0,1)$ to $(1,1)$. A classifier that guesses randomly is represented by a diagonal line from $(0,0)$ to $(1,1)$. Therefore, a perfect classification model has an AUC of 1.0. Likewise, a classifier that guesses randomly has an AUC of 0.5 \cite{auc_roc_admt}. Hence, the larger the AUC, the better the classifier performs on average. Furthermore, an important characteristic of the AUC is that it is invariant to the class prior probabilities \cite{auc_roc_pr}.

% It could be argued that an intention to speak is continuous, as sometimes an intention to speak is not as strong as another time. However, we make the assumption that intentions to speak are binary; someone wants to say something, or they do not. For the successful intention cases, the measures will be precision, recall and F1 score compared to the ground truth - being a turn start. For the unsuccessful intention cases, intention annotations are needed. Assuming that these are available, the same metrics can be used. For completeness, we use the following definitions from \cite{precision_recall}:

% \begin{equation}
%     \textrm{precision} = \frac{tp}{tp + \textit{fp}}
% \end{equation}
% \begin{equation}
%     \textrm{recall} = \frac{tp}{tp + \textit{fn}}
% \end{equation}
% \begin{equation}
%     \textrm{F1} = 2 * \frac{\textrm{precision} * \textrm{recall}}{\textrm{precision} + \textrm{recall}}
% \end{equation}

% Where \textit{tp} means the number of true positives, \textit{fp} means the number of false positives and \textit{fn} means the number of false negatives. We omitted the use of the accuracy measure because it can be misleading if the data is imbalanced. We believe that the data we will be working with is imbalanced because there will be many time points where someone does not have an intention to speak and only a few where there is an intention.

\section{Results}
\label{sec:res}
% You describe the results as they are, and how these related to the hypotheses (do they support the hypotheses or not)

\subsection{Insights from the Exploratory Study and Annotations}
From the exploratory study, we found a few useful insights. The first insight is that intentions to speak can be split up into intentions to \textit{start} speaking and intentions to \textit{continue} speaking. Also some cues that are likely indicative of intentions to speak were found. The most important one is the possibility to occasionally \textit{hear} mouth opening patterns. This can for example be heard by lip smacks or tongue clicks. In the annotation phase, this cue was perceived at least once for 7 out of the 13 people annotated. Hence, the insight of being able to occasionally hear mouth-opening patterns is likely useful for other research into intentions to speak as well and is not too person-specific. We interpret this as audible mouth-opening patterns, as investigated by \cite{mouth_open_patterns}. These sounds can be heard after applying a high-pass filter as well (see appendix \ref{app:smack}). Furthermore, it was found that people sometimes lean in towards someone's ear when they want to say something in the presence of outside noise. This is essentially a posture shift, but this specific posture shift is worth highlighting. Lastly, throat clears are sometimes perceived as a turn-initializing signal.

\subsection{Quantitative Evaluation of the Model}
\label{sec:quantitative}
The models are evaluated by obtaining AUC scores under different lengths of the time window. The different time windows used are 1, 2, 3, and 4 seconds. The model is trained on all data except second 3600 until 4200 because this segment was annotated and is used as the test set. For model validation, 3-fold cross-validation is used to optimize the model parameters and training stops after 10 epochs. The model is evaluated in five experiments: prediction of all\footnote{``All here means the union of extracted successful cases and annotated perceived unsuccessful cases"} intentions to speak, prediction of successful intentions to speak, prediction of all annotated unsuccessful intentions to speak, prediction of unsuccessful intentions to start speaking and prediction of unsuccessful intentions to continue speaking.

In the first experiment, the positive samples are the accelerometer data corresponding to the time window before the participant speaks (see \ref{sec:successful__case_extraction}), and the annotated perceived unsuccessful intentions to speak. In the second experiment, the positive samples are only the accelerometer data corresponding to the time window before the participant speaks. Strictly speaking, this is the task the model was trained for. In the third experiment, the positive samples are the accelerometer data corresponding to the time intervals that were manually annotated as perceived intentions to speak. The end time of these positive samples is taken from the annotation and the start time of the positive samples depends on the experimental condition (between 1 to 4 seconds before the end time). In the fourth experiment, the positive samples are only the accelerometer data corresponding to the time intervals that were annotated as ``INTS\_start". Similarly, in the fifth experiment, the data is generated the same as in the fourth experiment. The only difference is that the labelled ``INTS\_continue'' cases are used instead of ``INTS\_start". The positive samples are left out if they overlap with speech. Furthermore, the negative samples for each experiment are randomly sampled accelerometer data without overlap with the respective positive samples. More specifically, for the unsuccessful cases, negative samples do not only exclude overlap with the corresponding unsuccessful positive samples, but also exclude overlap with positive cases.

\begin{figure}[H]
\centering
  \includegraphics[scale=0.5]{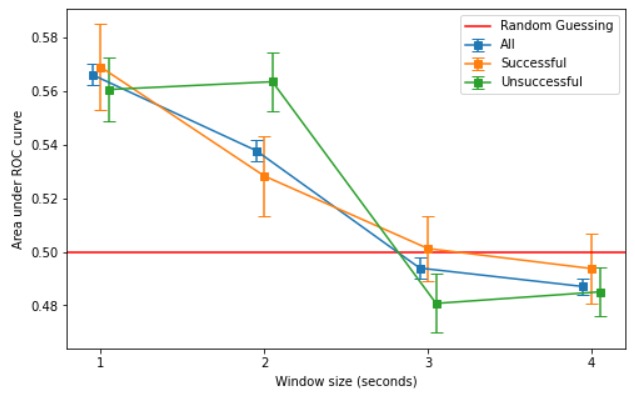}
  \caption{Visualisation of the first three rows of table \ref{tab:auc_results}. Note that points are shifted slightly left or right to prevent overlap of the standard deviations.}
  \label{fig:plot_all_suc_unsuc}
\end{figure}

\begin{figure}[H]
\centering
  \includegraphics[scale=0.5]{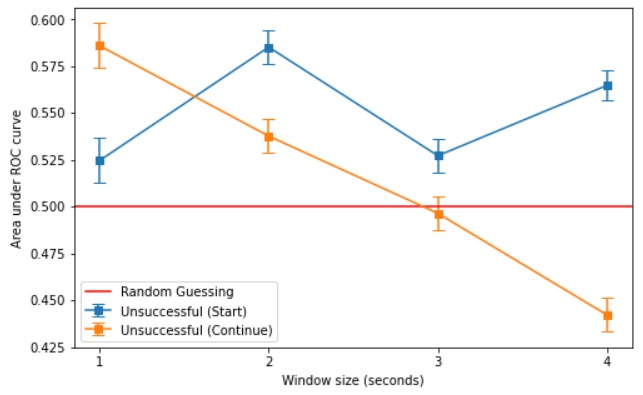}
  \caption{Visualisation of the final two rows of table \ref{tab:auc_results}.}
  \label{fig:plot_start_cont}
\end{figure}

For all experiments, the trained models were used to analyze the classification quality on the test sets under different time windows. This procedure was repeated 100 times to make the results more stable and to get a reliable estimate of the means and standard deviations of the area under the ROC curve. For the results, see table \ref{tab:auc_results}. These results are visualized in Figures \ref{fig:plot_all_suc_unsuc} and \ref{fig:plot_start_cont}.

\begin{table}[H]
\centering
\begin{tabular}{|l|l|l|l|l|}
\hline
AUC scores       & 1 second  & 2 seconds & 3 seconds & 4 seconds \\ \hline
All intentions to speak & 0.5661 (0.004) & 0.5378 (0.004) & 0.4939 (0.004) & 0.4870 (0.003)\\ \hline
Successful  & 0.5690 (0.016) & 0.5283 (0.015)  & 0.5012 (0.012)  & 0.4937 (0.013) \\ \hline
Unsuccessful & 0.5606 (0.012) & 0.5635 (0.011) & 0.4807 (0.011) & 0.4850 (0.009)\\ \hline
Unsuccessful (Start) & 0.5246 (0.012) & 0.5852 (0.009) & 0.5272 (0.009) & 0.5648 (0.008) \\ \hline
Unsuccessful (Continue) & 0.5860 (0.012) & 0.5377 (0.009) & 0.4964 (0.009) & 0.4421 (0.009)\\ \hline

\end{tabular}
\caption{Mean and standard deviation of ROC AUC scores for the evaluation on successful intentions to speak, unsuccessful intentions to start speaking and unsuccessful intentions to continue speaking. The experiments are done for four different window sizes.}
\label{tab:auc_results}
\end{table}

% \begin{figure}[h]
% \centering
%   \includegraphics[scale=0.35]{Figure_1.png}
%   \caption{Mean and standard deviation of ROC AUC scores for the evaluation on successful intentions to speak, unsuccessful intentions to start speaking and unsuccessful intentions to continue speaking. The experiments are done for four different window sizes.}
%   \label{fig:Comparison of successful and unsuccessful case}
% \end{figure}

%caption: Mean and standard deviation of ROC AUC scores for the evaluation on successful intentions to speak, unsuccessful intentions to start speaking and unsuccessful intentions to continue speaking. The exeriments are done for four different window sizes.

\subsection{Statistical Analysis}
To analyse the significance of the results, some statistical tests are done. Section \ref{sec:ttest} tests the hypotheses presented in \ref{sec:hypotheses}. The other subsections are added to explore the results more in-depth.

\subsubsection{Comparison to Random Guessing}
\label{sec:ttest}
To test the hypotheses, t-tests are done. The null hypothesis is $H_0$: ``The model performs the same as or worse than random guessing". The alternative hypothesis is $H_1$: ``The model performs better than random guessing". For each model evaluation configuration, as in table \ref{tab:auc_results}, a mean and a standard deviation of the AUC scores exist. As explained in section \ref{sec:measures}, random guessing has a mean AUC of 0.5. In addition, we assume that the standard deviations of the AUC for random guessing are the same as the standard deviations in table \ref{tab:auc_results}. One-sided t-tests are done to see if the performance is better than random guessing. The p-values of the t-tests are given in table \ref{tab:ttest}. The values are compared to the (conservative) threshold of 0.001 and the comparison results are indicated with color in the table. Green color means that the result is significant and the null hypothesis can be rejected, meaning that the model performed better than random guessing. Red color means that there is not enough evidence to reject the null hypothesis.

\begin{table}[h]
\centering
\begin{tabular}{|l|l|l|l|l|}

\hline
p-values       & 1 second  & 2 seconds & 3 seconds & 4 seconds \\ \hline
All intentions to speak &\cellcolor{sig} 6.545e-185 &\cellcolor{sig} 4.287e-138 &\cellcolor{nsig} 1.0000 &\cellcolor{nsig} 1.0000\\ \hline
Successful  &\cellcolor{sig} 4.889e-77 &\cellcolor{sig} 1.099e-29  &\cellcolor{nsig} 0.2402  &\cellcolor{nsig} 0.9996 \\ \hline
Unsuccessful &\cellcolor{sig} 1.575e-88 &\cellcolor{sig} 1.166e-98 &\cellcolor{nsig} 1.0000 &\cellcolor{nsig} 1.0000\\ \hline
Unsuccessful (Start) &\cellcolor{sig} 3.130e-33 &\cellcolor{sig} 3.069e-138 &\cellcolor{sig} 1.293e-53 &\cellcolor{sig} 1.725e-125 \\ \hline
Unsuccessful (Continue) &\cellcolor{sig} 1.241e-115 &\cellcolor{sig} 5.528e-75 &\cellcolor{nsig} 0.9974 &\cellcolor{nsig} 1.0000\\ \hline

\end{tabular}
\caption{P-values for the t-tests comparing the model performance to random guessing. Green indicates that the result is significant, red color indicates that the result is not significant enough to reject the null hypothesis.}
\label{tab:ttest}
\end{table}

\subsubsection{Linear Regression Through AUC Scores}
Many of the data presented in Figures \ref{fig:plot_all_suc_unsuc} and \ref{fig:plot_start_cont} indicate downwards trends. To explore this visual intuition statistically, linear regression is applied. The slopes found by linear regression through the results of each experiment are presented in Table \ref{tab:regress}. Additionally, the $R^2$ measure (coefficient of determination) is presented. This measure indicates how well the regression model fits the data.
% \begin{table}[h]
% \centering
% \begin{tabular}{|l|l|l|l|}

% \hline
%                         & Slope  & p-value & $R^2$ \\ \hline
% All intentions to speak & -0.0281 &\cellcolor{sig} 3.566e-228 & 0.927 \\ \hline
% Successful              & -0.0253 &\cellcolor{sig}  6.006e-122 & 0.750 \\ \hline
% Unsuccessful            & -0.0309 &\cellcolor{sig} 2.468e-109 & 0.711 \\ \hline
% Unsuccessful (Start)    & 0.0063 &\cellcolor{sig} 1.971e-07 & 0.0658\\ \hline
% Unsuccessful (Continue) & -0.0473 &\cellcolor{sig} 1.470e-296 & 0.967 \\ \hline

% \end{tabular}
% \caption{TODO}
% \label{tab:regress}
% \end{table}

\begin{table}[h]
\centering
\begin{tabular}{|l|l|l|}

\hline
                        & Slope & $R^2$ \\ \hline
All intentions to speak & -0.0281 & 0.927 \\ \hline
Successful              & -0.0253 & 0.750 \\ \hline
Unsuccessful            & -0.0309 & 0.711 \\ \hline
Unsuccessful (Start)    & 0.0063 & 0.0658\\ \hline
Unsuccessful (Continue) & -0.0473 & 0.967 \\ \hline

\end{tabular}
\caption{Slopes and $R^2$ values for linear regressions through the results. $R^2$ close to 0 indicates a bad fit and close to 1 indicates a good fit.}
\label{tab:regress}
\end{table}

\subsubsection{Difference Between Start and Continue}
The data visualized in Figure \ref{fig:plot_start_cont} might raise the question if there is a significant difference between the two cases. For every window size, a Welch's t-test\footnote{Does not necessarily assume equal variance for the two samples.} is performed, comparing the data. The null hypothesis is $H_0$: ``The samples have the same mean". The alternative hypothesis is $H_1$: ``The samples have different means".
The p-values for a window size of 1, 2, 3, and 4 seconds are 1.979e-89, 4.843e-93, 5.594e-62, and 3.325e-175 respectively. Hence, for each considered window size, the null hypothesis can safely be rejected. 

\subsection{Discussion}
% You discuss the meaning of the results in more depth also in relation to the work of others introduced in 1.1
% MOST IMPORTANTLY: INTERPRET THE RESULTS
From the results of the experiments, it is clear that for nearly all cases, the mean AUC decreases as the time window increases. One exception is that for the unsuccessful ``INTS\_start" cases, there is no obvious tendency for the AUC to increase or decrease as the time window increases. The corresponding $R^2$ value from table \ref{tab:regress} also clearly indicates this. It can also be observed that in the successful intention cases, most of the accelerometer information related to speech intention is concentrated in a very small period before the speaking starts.
We speculate that these are patterns in the breathing of the participant, as in \cite{wlodarczak2020breathing}. They might also be patterns in posture shifts, as in \cite{whosnext}.
The concentration of the features into the small period before speaking may be because the participant thinks they have a high probability of successfully speaking, and manifest their intention less. As for the unsuccessful ``INTS\_start" cases, the reason why effective information does not decrease with the increase of the window size may be because the participants feel that they have a very small chance of speaking, so the motivation to speak is manifested stronger. They still try to speak in turn-taking through behaviours that will bring effective accelerometer information for a longer period compared to the successful intention case.

Obviously, the data close to the start of speech is still included in the longer time window. Yet, more data is added. Therefore, we speculate that this data is often less informative, making it harder for the model to pick up on the useful patterns that are close to speech.

Another trend is that the mean AUC under the unsuccessful ``INTS\_continue" condition decreases as the time window increases. This may be because the labeling conditions compared to the ``INTS\_start" cases are different. For the ``INTS\_continue" cases, the participant already had the turn before but was for example interrupted. Hence, the cues to indicate an intention to speak are not as long-lasting as in the ``INTS\_start" condition because they were already speaking, and therefore did not need to show social cues that indicate the intention to speak. The explanation for this insight is speculative because there are too few manually labelled samples under the ``continue" condition. However, with more labelled data, it may be possible to find deeper relationships between the degree of speaking intention and whether or not a participant had the turn before.

% Result of T test
Additionally, from our t-test results, it can be seen that using accelerometer data to train the model on successful intention cases can indeed predict speech intent with a significant advantage over random guessing using the 1s and 2s time window. Moreover, the model's prediction results for unsuccessful intention to start speaking under the 3s and 4s time windows are also significantly better than random guessing. The speculative explanation for this result is consistent with the explanations for previous experiment results, possibly because in these cases, the participant's intention to speak is the strongest. Therefore, in the conditions of the 3s and 4s window, this intent has not yet significantly declined and the added data is still informative.

% limitation of negative samples
There are two limitations in the experiments. The first is the small amount of annotated samples; this might have played a role in the results. The second limitation is that in the sampling strategy for negative samples, we randomly sample periods that do not contain positive samples. Therefore, negative samples may partially overlap with the periods when the participant is speaking, which means that such negative samples can contain some accelerometer data related to speech. They may also not overlap at all with the segments when the participant is speaking, meaning that the negative sample's accelerometer data is related to the state of the participant when they are completely silent. Since the accelerometer information in the speaking state may be different from that in the silent state, the model might have picked up on patterns in this, rather than in fact recognizing intentions to speak. The proportion of overlap negative samples with silence and speech could be found in \ref{app:negative sample proportion}. Additionally, positive samples may also be related to the length of time since the participant's last speech. These assumptions are highly worth exploring in future work.

Finally, it is worth stressing again that randomly guessing would yield an AUC score of 0.5 on average \cite{auc_roc_admt}. Most of the results found have scores above 0.5. However, the scores are often not much higher than this. This would indicate that the model has picked up on some patterns and has \textit{some} predictive power, but is also not very reliable in terms of precision and recall. There are a few possible explanations for this. First of all, there is likely a natural limit to the AUC score in this setting. For example, posture shifts are correlated with turn-taking \cite{whosnext}, but people also often shift posture without having an intention to speak or have an intention to speak without shifting their posture. Secondly, the physical movement information related to speaking intent may sometimes be too subtle, and it may be difficult to detect e.g. deep breaths (which might be useful according to \cite{wlodarczak2020breathing}) if the sensor is not sensitive enough. Thirdly, the accelerometer data that can be collected may be affected by the placement of the sensor and the degree of interaction between other body movements, such as head or hand movements that do not involve the trunk.

\section{Conclusion}
This project focuses on predicting the intentions to speak using in-the-wild accelerometer data. Speaking intentions are split into two cases: successful and unsuccessful cases. A subset of perceived unsuccessful intentions is annotated manually. These unsuccessful intentions are divided into intentions to start speaking, and intentions to continue speaking. Following the experiment results, we conclude with answers to our research questions. Specifically for a window size of 1 or 2 seconds, using accelerometer data as input, the neural network model predicts the successful and unsuccessful intentions to speak better than random guessing. A significant difference between the performance of different time windows can be noticed. An observed trend seems to be that the AUC score decreases as the window size increases, with perceived unsuccessful intentions to start speaking being an exception.

In the exploratory study of the REWIND dataset, we found that tongue clicks and lip smacks can be heard as an indication of an intention to speak. We interpret this as audible mouth-opening patterns, as investigated by \cite{mouth_open_patterns}. These sounds can be heard under a high-pass filter as well (see appendix \ref{app:smack}). In this work, the insight of audible mouth-opening patterns is only used for the annotations, but it might be useful for future work in the direction of inferring intentions to speak. To answer the research questions more in-depth and to investigate other directions of the same problem, possible future work will be discussed in the next section.
% Moreover, a specific type of posture shift is also found as a social cue for the intention to speak, which is leaning towards the listener's ear when the environment is noisy. 

% 1. found the sound of tongue clicks and lip smacks
% 2. model performs better than random guessing
% 3. increase the time window, auc score of all intentions decreases
% 4. increase the time window, auc score of unsuccessful continue decreases

\section{Further Research}
\label{sec:future}
% Your final conclusion and recommendations for further/future work
Based on the limitations of our experiments and conclusions, we propose some possibilities for future work. First, due to the reality of limited time and resources, we only annotated a subset of perceived unsuccessful intentions to speak. However, more annotations of the unsuccessful intentions to speak can always help to test the model and improve its robustness. Once there are plenty of annotations of unsuccessful intentions, it would also be feasible to train the model only on those annotations of unsuccessful intentions to speak. In this way, it will only focus on those unrealised intentions. Secondly, our experiment trains the model with successful intentions to speak, but it would be interesting to train the model on both successful intentions and the annotated unsuccessful intentions to speak, as they are both considered intentions to speak. Further research on these points would answer the ``to what extent"-part of the main research question more in-depth. Moreover, the model itself can be further improved as well. In our experiment, we tested for different sizes of time windows and the number of epochs but did not optimize all other hyperparameters of the model. Thus another possible improvement would be optimizing the architecture and the hyperparameter of the model. 

The model we refactored from is taking a multi-modal fusion as input, due to time limitations and consideration of privacy, we only use accelerometer data. Future research might look into hallucinating higher-frequency accelerometer data from low-frequency data. Furthermore, as previously discussed, including more modalities such as audio and video is likely to improve the performance of the model. It is possible to use high-frequency audio to detect the smacks and tongue clicks\footnote{For some initial experiments in this direction, see appendix \ref{app:smack}}. Besides, as mentioned before, posture shifts and gestures are correlated with speaking intentions, but those posture shifts detected from accelerometer data do not capture intentions to speak very accurately. Hence, it is worth investigating if directly extracting poses from video data would improve the performance. This kind of data would also capture patterns in hand gestures and head movements. Some research would also need to look into how privacy-preserving high-frequency audio and extracted postures really are. 

There are some other considerations of this project. Currently, the data with successful intentions is generated from VAD files, but this can still include some flaws. The parameters for the successful case extraction could be chosen more robustly; as of now, they are manually optimized and empirically checked for quality. Speaker diarization literature might contain useful insights on how to do this processing step better. An ideal setting would be annotating the successful turn-changes manually, which excludes the effect of incorrectly filtered pauses or back-channelling. Another consideration is related to the granularity of the time window. During the experiment, we choose a granularity of 1 second, which is based on the synchronized level of the REWIND dataset. However, detectable intentions could appear within a very small time interval. Smaller scales of granularity also contribute to finding a more precise optimal time window. Thus, it would be interesting to test the model with smaller scales of granularity. 

A final consideration is about differences in social cues across cultures. The dataset used in this project is fully set in Dutch, which means participants are Dutch speakers with similar backgrounds. However, people with different cultural backgrounds tend to behave with different social cues, and they also behave differently when it comes to multi-culture communication. Hence, the cultural background needs to be considered as a factor that could influence the dataset and the annotations. Executing research similar to this, but using data and annotators from countries other than the Netherlands, and inspecting the differences might be interesting future work.

% missed cases in annotation?

% Future work ideas:
% 1. More annotations of unsuccessful INTS
% 2. Train model on annotations + successful INTS
% 3. Train model on only annotations of unsuccessful INTS (there is not much of a point in predicting the successful cases anyway)
% 4. Optimizing model architecture/hyperparameters more
% 5. Maybe annotate turns rather than using VAD
% 6. Use extracted pose
% 7. Use high-frequency audio (maybe smack-detector)

% Other stuff to discuss:
% 1. Maybe annotate turns rather than using VAD (pause/backchannel filtering is never 100% correct)
% 2. Difference in social cues across cultures? (this is a Dutch setting)
% 3. Not enough annotated data points maybe

\subsubsection*{Acknowledgements}
We would like to thank Hayley Hung and José Vargas Quiros for supervising this project and for providing valuable feedback. Next, we would like to thank the other staff and fellow students of the Social Signal Processing course for posing interesting questions and providing feedback after the intermediary presentations. Finally, we would like to thank Ruud de Jong for providing access to the soundproof booth in the INSYGHTLab.

\printbibliography

\newpage
\appendix
\section{Dataset Considerations}
\label{data-considerations}
In the initial phase of the research, a decision had to be made on what dataset should be used for this research. This section described the considerations taken into account for the datasets: Conflab, REWIND, Dominos and MULAI.

The general settings for the four candidate datasets are as follows. Conflab has recorded a real social event at the 2019 ACM Multimedia international conference, including 48 participants and the corresponding combination of social interaction features, capturing in-the-wild free-standing social conversations. REWIND collects the data of an indoor crowded social event when standing people are free to walk and talk to others. For both Conflab and REWIND, video data is recorded by overhead cameras, and individuals have wearable accelerometers. Additionally, REWIND has elevated side-view cameras and wearable per-person microphones. The data collection scenario for the Dominos dataset is that three people are sitting together in a room, and they are talking to each other while working on a domino task. Finally, the MULAI corpus is a dataset that records dyadic human-human interactions in a room, aiming to understand the correlation between acoustic laughter properties and annotated humour ratings.

\begin{figure}[ht]
\centering
\includegraphics[width=\textwidth]{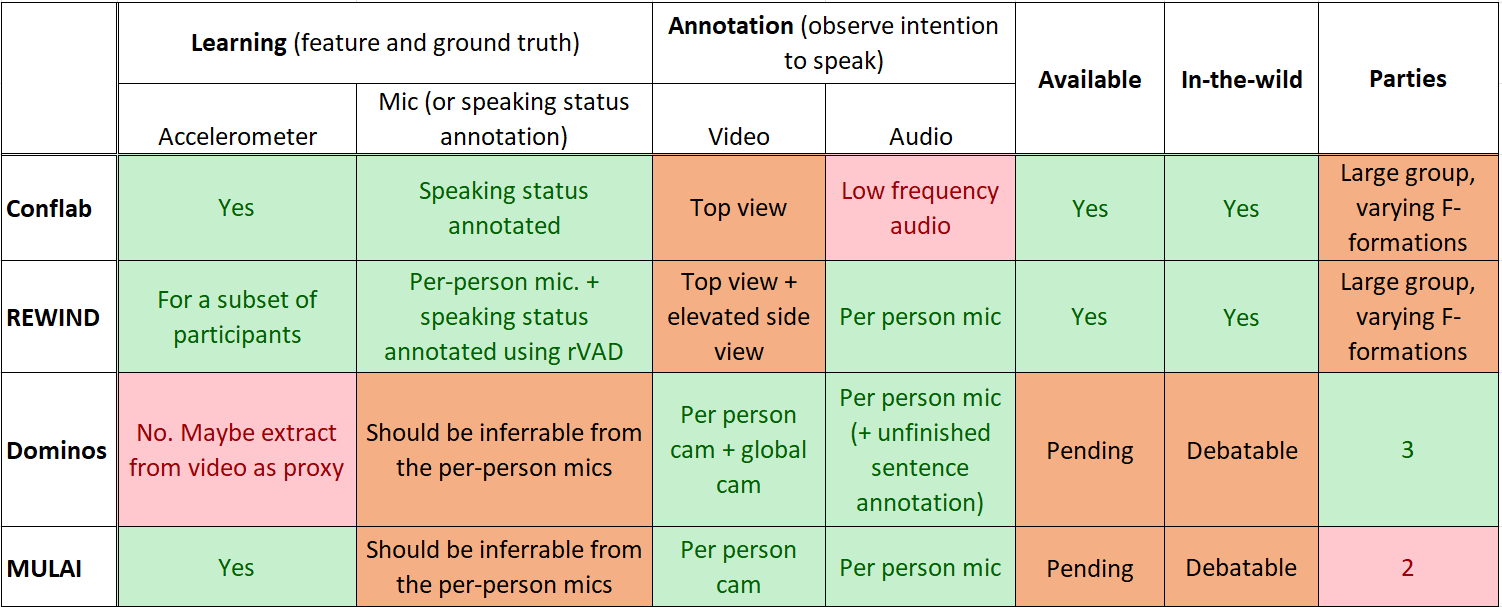}
\caption{Datasets Comparison}
\label{fig:Datasets Comparison}
\end{figure}

In table \ref{fig:Datasets Comparison}, the four candidate datasets have been compared. Conflab and MULAI all have the accelerometer feature while only a subset of the participants in REWIND has this feature, and dominos doesn't have a specific accelerometer feature but in theory, it could be extracted from the video (pose estimation) as a proxy. And both conflab and REWIND have annotations for speaking status, while these have to be inferred from the mics in Dominos and MULAI. This could be a complicating factor due to possible noise in the audio. 
For the manual annotation of perceived intentions, the camera and audio setup and quality are important. A per-person camera and microphone would be ideal, as the people can be closely inspected to see any cue that indicates an intention to speak. The most ideal video and audio setups are the Dominos and MULAI datasets since there are cameras and microphones for each person. Dominos even has an additional unfinished sentence annotation, which can be useful for identifying intentions to speak. Meanwhile, there are only top views available for the groups of people in Conflab and REWIND, but REWIND has some additional elevated side views. Moreover, the audio of Conflab is low frequency for privacy concerns. This will make it difficult to observe intentions to speak. Conflab and REWIND are already available for the project, but access to the Dominos and MULAI datasets is still pending. Furthermore, the data collection scenarios for Conflab and REWIND are in-the-wild, and the scenarios for Dominos and MULAI are debatable since they are in a lab setting. In addition, different numbers of people appear in the interaction scenes of Conflab and REWIND. Meanwhile, there are only 3 participants and 2 participants for Dominos and MULAI respectively. On the one hand, the varying group size could be a complicating factor (extra degree of freedom), whereas a constant number of interaction participants can take this degree of freedom away. On the other hand, this varying group size could make the approach more ubiquitous. Finally, the dyadic interactions in MULAI are maybe not the best, as the focus is on multi-party interactions.
 
\section{Samples and Speech Overlap}
\label{app:negative sample proportion}
\begin{figure}[H]
\centering
\includegraphics[width=\textwidth]{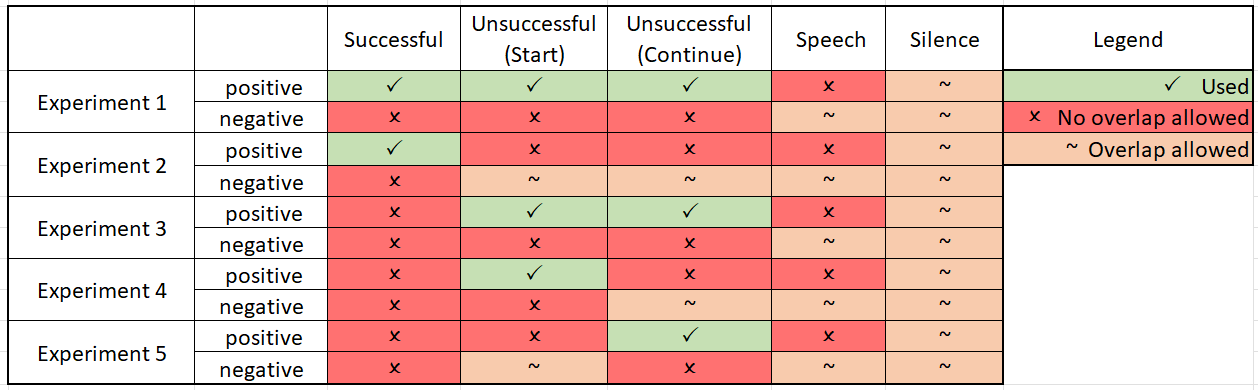}
\caption{Samples for different experiments. Experiment numbers are consistent with the order in table \ref{tab:auc_results} and \ref{tab:ttest}.}
\label{fig:Samples table}
\end{figure}

\begin{figure}[H]
\centering
\includegraphics[width=\textwidth]{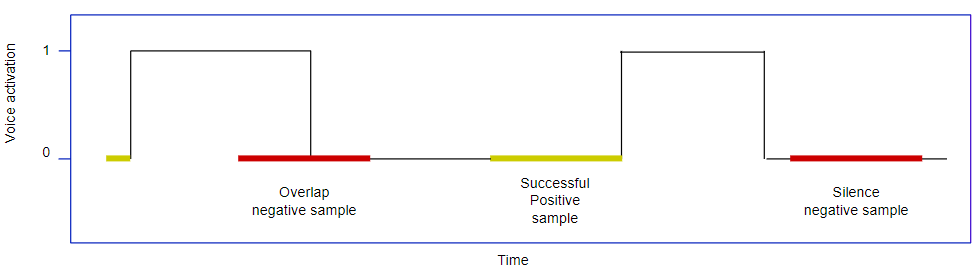}
\caption{Examples of positive and negative samples in the successful intention case.}
\label{fig:Successful samples}
\end{figure}

\begin{figure}[H]
\centering
\includegraphics[width=\textwidth]{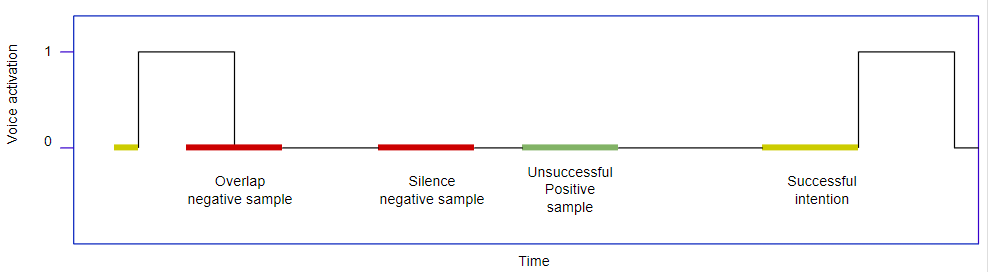}
\caption{Examples of positive and negative samples in the unsuccessful intention case.}
\label{fig:Unsuccessful samples}
\end{figure}

\begin{figure}[H]
\centering
\includegraphics[width=\textwidth]{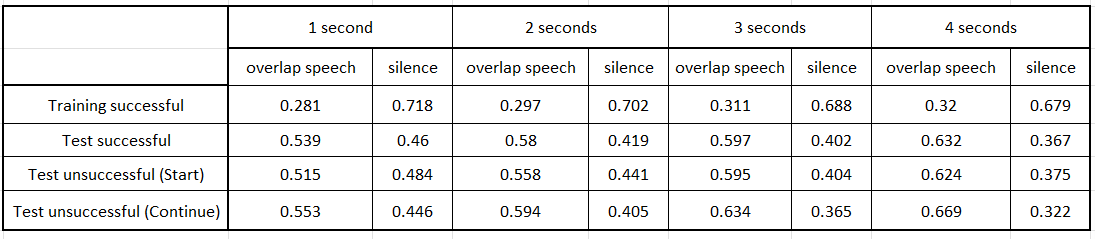}
\caption{Proportions of negative samples that either have overlap with silence or speech.}
\label{fig:Proportion}
\end{figure}

\begin{figure}[h]
\centering
\includegraphics[width=\textwidth]{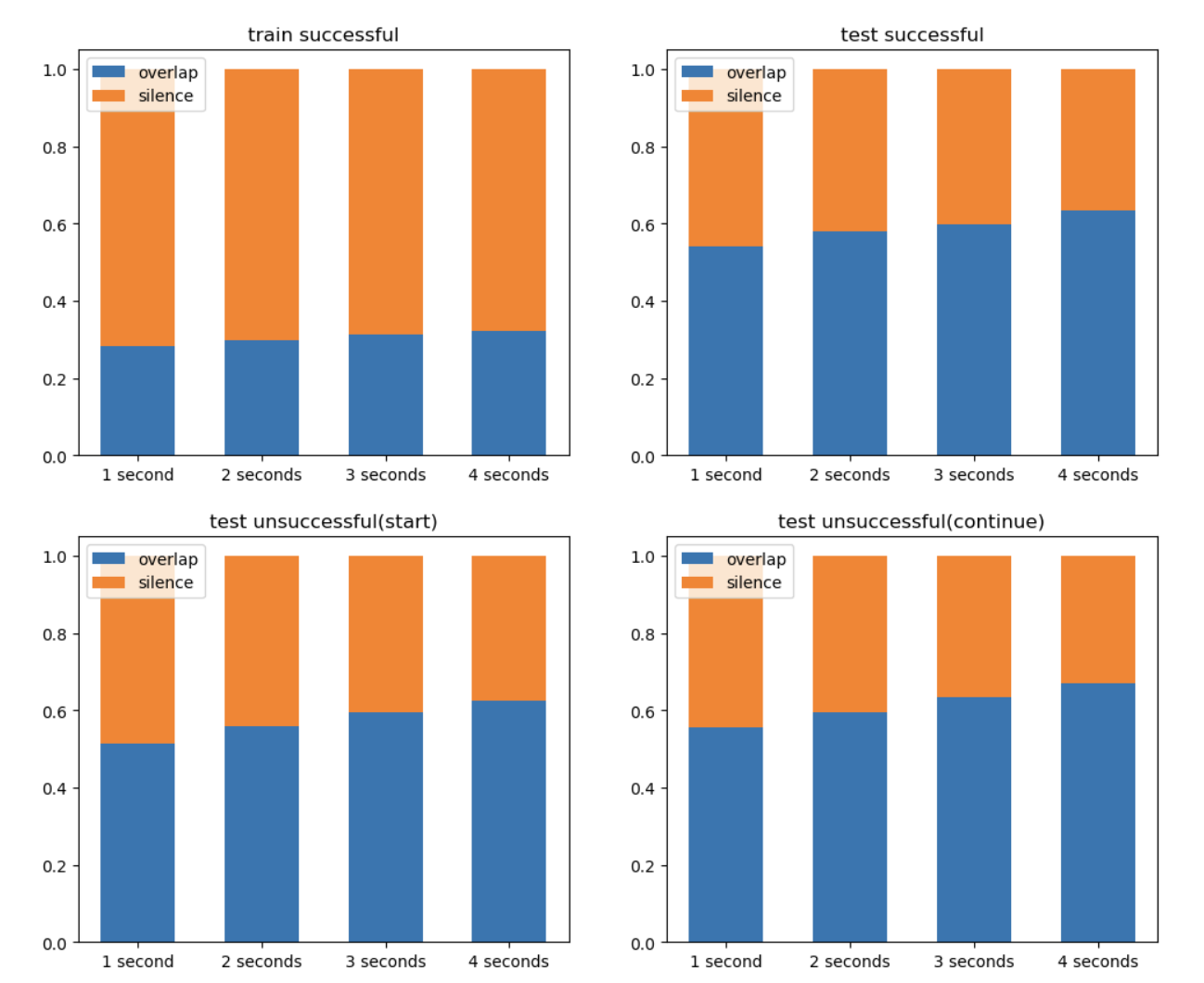}
\caption{Proportions of negative samples that overlap with speech. Visualization of Figure \ref{fig:Proportion}.}
\label{fig:Proportion_bar}
\end{figure}

The positive samples are the accelerometer data corresponding to the time window before the participant speaks. On the other hand, negative samples are randomly sampled during periods that do not include intention (not including successful intention in successful intention cases, and not including both successful and unsuccessful intentions in unsuccessful cases). What counts as positive and negative samples in different experiments is shown as a diagram in Figure \ref{fig:Samples table}. An example of how positive and negative samples are generated can be found in Figure \ref{fig:Successful samples} and Figure \ref{fig:Unsuccessful samples}. Under this sampling strategy, negative samples can further be distinguished as 1. Negative samples that have (partial) overlap with the participant's own speaking period. 2. Negative sample periods that do not overlap with the participant's own speaking period, indicating that the participant is silent. These proportions can be found in Figure \ref{fig:Proportion} and are visualized in \ref{fig:Proportion_bar}. 

\newpage
\section{Audible Mouth-Opening Patterns}
\label{app:smack}
\begin{figure}[h]
\centering
\includegraphics[width=\textwidth]{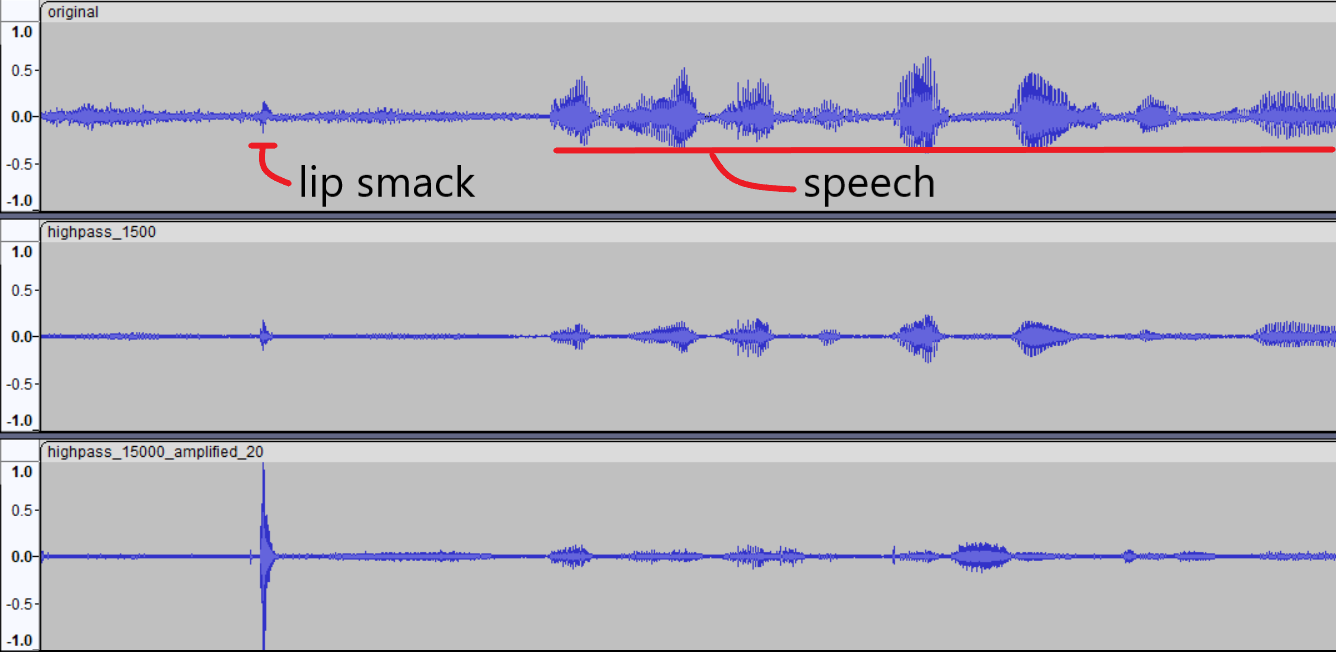}
\caption{Visualisation of three audio files: original audio, audio high-pass filtered with a frequency of 1500Hz, and audio high-pass filtered with a frequency of 15000Hz and amplified by 20 times.}
\label{fig:smack}
\end{figure}

To further investigate the possible use of audible mouth-opening patterns, a simple experiment was done. An audio sample was taken where a participant makes a lip-smacking sound preceding her speech. Two processed versions of this signal are generated; one where only the audio waves with a higher frequency than 1500Hz are kept, and one where only the audio waves with a frequency higher than 15000Hz are kept. This latter signal is then also amplified by 20 times. Figure \ref{fig:smack} shows three visualisations of audio files: the original sound and the two processed versions respectively. Listening to especially the second processed version, the smacking sound is easily audible as a loud click, but the person's voice is distorted in such a way that it would be difficult to identify the person based on their voice.

Additionally, some experiments were done using a low-pass filter, but with this, it seemed that the smacking sound disappeared. High-frequency audio might be useful for future work for the privacy-preserving inference of intentions to speak. Though, additional work would need to look into hallucinating low-frequency audio from high-frequency audio to verify how privacy-preserving it really is. 

In the annotation phase, audible mouth-opening patterns were perceived at least once for 7 out of the 13 people annotated. Hence, the insight of being able to occasionally hear mouth-opening patterns is likely useful for other research into intentions to speak as well and is not too person-specific. It was sometimes found difficult to identify which specific mouth-opening pattern was heard (e.g. tongue click or lip smack). Hence, only a broader categorization was used in the annotation phase. Perhaps individual categories correlate differently with intentions to speak. Investigating this is interesting future work.

\end{document}